\documentclass{article} 
\usepackage{algorithm2e}
\usepackage{graphicx}

\usepackage{caption,booktabs}
\usepackage{subcaption}
\usepackage{graphicx}
\usepackage{cite}
\usepackage{amscd,amssymb} 
 \usepackage{array}
\usepackage{xy} 
\usepackage{amsmath} 
\usepackage{fancyhdr}
\usepackage{setspace} 
\usepackage{verbatim} \usepackage{multicol} \usepackage{longtable}
\usepackage{amsthm}
\newtheorem{thm}{Theorem}[section]

\theoremstyle{remark}
\newtheorem{rem}[thm]{Remark}

\newcommand{\norm}[1]{\left\Vert#1\right\Vert}
\newcommand{\abs}[1]{\left\vert#1\right\vert}

\newcommand{\Real}{\mathbb R}

\newcommand{\vc}[1]{#1}

\renewcommand{\d}[1]{\text{ d}#1}

\newcommand{\GP}{\mathcal {GP} }

\newcommand{\expect}[1]{\ensuremath{ \langle#1\rangle } }

\newcommand{\var}[1]{\ensuremath{ \text{var}[#1]} }

\newcommand{\rank}{\ensuremath{ \text{ rank }}}

\newcommand{\state}{\ensuremath{ \vc x}} 
\newcommand{\statespace}{\ensuremath{ \mathcal X}}
\newcommand{\configspace}{\ensuremath{ \mathcal Q}}

\newcommand{\ctrlspace}{\ensuremath{ \mathcal U}}

\renewcommand{\d}{\ensuremath{\text{ d}}}

\newcommand{\beq}{\begin{equation}}
\newcommand{\eeq}{\end{equation}}

\title{Stochastic processes and feedback-linearisation for online identification and Bayesian adaptive control of fully-actuated mechanical systems}

\author{
Jan-P. Calliess, Antonis Papachristodoulou and Stephen J. Roberts   \\
Department of Engineering Science, 
Oxford University, 
UK \\
}

\begin{document}

\maketitle

\begin{abstract}
\begin{quote}
This work proposes a new method for simultaneous probabilistic identification and control of an observable, fully-actuated mechanical system. 
Identification is achieved by conditioning stochastic process priors on observations of configurations and noisy estimates of configuration derivatives. In contrast to previous work that has used stochastic processes for identification, we leverage the structural knowledge afforded by Lagrangian mechanics and learn the drift and control input matrix functions of the control-affine system separately.  
We utilise feedback-linearisation to reduce, in expectation, the uncertain nonlinear control problem  to one that is easy to regulate in a desired manner. Thereby, our method combines the flexibility of nonparametric Bayesian learning with epistemological guarantees on the expected closed-loop trajectory. We illustrate our method in the context of torque-actuated pendula where the dynamics are learned with a combination of normal and log-normal processes.
\end{quote}
\end{abstract}

\section{Introduction}

Control may be regarded as decision making in a dynamic environment. Decisions have to be based on beliefs over the consequences of actions encoded by a model. Dealing with uncertain or changing dynamics is the realm of adaptive control. In its classical form, parametric approaches are considered (e.g. \cite{4739101} ) and, typically, uncertainties are modelled by Brownian motion (yielding stochastic adaptive control \cite{kumar1985,duncan:08}) or via set-based considerations (an approach followed by robust adaptive control \cite{robustadaptctrlbook:95}). In contrast, we adopt an epistemological take on probabilistic control and bring to bear Bayesian nonparametric learning methods whose introspective qualities \cite{Posnerintrospection:2013} can aide in addressing theexploration-exploitation trade-offs  relative to one's subjective beliefs in a principled manner\cite{Alpcan2011}. Based on these Bayesian learning methods, it is our ambition to develop adaptive controllers with probabilistic guarantees (interpreted in an epistemological sense) on control success.
%

In contrast to classical adaptive control where inference has to be restricted to finite-dimensional parameter space, the nonparametric approach affords the learning algorithms with greater flexibility to identify and control systems with very few model assumptions. This is possible because these methods grant the flexibility to perform Bayesian inference over rich, infinite-dimensional function spaces that could encode the dynamics. This property has led to a surge of interest in Bayesian nonparametrics; particularly benefiting their algorithmic advancement and application to a plethora of learning problems. Due to their favourable analytic properties, \textit{normal} or \textit{Gaussian} processes (GPs) \cite{bauer2001, GPbook:2006} have been the main choice of method in recent years. Among other domains, GPs have been applied to learning discrete-time dynamic systems in the context of model-predictive control \cite{Kocijan2005,Kocijan2003,Murray-Smith2003,Rogers2011}, learning the error of inverse kinematics models \cite{tuongpetersICRA:2010, Nguyen-Tuong2008}, dual control \cite{Alpcan2011} as well as reinforcement learning and dynamic programming \cite{Deisenroth2008,Deisenroth2009,rottmannburgard:09GPvalit,KofoxhaehnelblimpctrlGP:07}.

On the flip side, the extent of flexibility can lead to the temptation to use the approach in a black-box fashion, disregarding most structural knowledge of the underlying dynamics \cite{Kocijan2005,Kocijan2003,rottmannburgard:09GPvalit,KofoxhaehnelblimpctrlGP:07,Murray-Smith2003}. This can result in unnecessarily high-dimensional learning problems, slow convergence rates and often necessitates large training corpora, typically to be collected offline. In the extreme, the latter requirement can cause slow prediction and conditioning times. Moreover, they have been used in combination with computationally intensive planning methods such as dynamic programming \cite{Deisenroth2008,Deisenroth2009,rottmannburgard:09GPvalit} rendering real-time applicability difficult. 

In contrast to all this work, we will incorporate structural a-priori knowledge of the dynamics afforded by Lagrangian mechanics (without sacrificing the  flexibility afforded by the nonparametric nature). This requires, in some instances, partial departure from Gaussianity (e.g. if the sign of a function component of the dynamics is known) but improves the detail with which the system is identified and can reduce the dimensionality of the identification problem. Furthermore, our method will use the uncertainties of the models to decide upon training example incorporation and decision making. 

Aside from learning, our method employs feedback-linearisation \cite{Spong94:PFL} in an outer-loop control law to reduce the complexity of the control problem.  Thereby, in expectation, the problem is reduced to controlling a double-integrator via an inner-loop control law. If we combine the outer-loop controller with an inner-loop controller that has desirable guarantees (e.g. stability) for the double-integrator, these properties can extend to the expected given non-linear closed-loop dynamics. The resulting approach enables rapid decision making and can be deployed online. 

Our work is presented at the AMLSC Workshop at NIPS, 2013. During the review process, we were made aware of GP-MRAC \cite{Chowdhary2013}. The authors utilise a Gaussian process on joint state-control space to learn the error of an inversion controller in model-reference adaptive control. Under the assumption that the GP could be stated as an SDE of time, they prove stability. 
In contrast to this work, our method is capable of identifying the drift and control input vector fields constituting the underlying control-affine system individually, yielding a more fine-grained identification result. While this benefit requires the introduction of probing signals to the control during online learning, each of the coupled learning problems has state space dimensionality only. Moreover, our method and stability results are not limited to Gaussian processes. If the control-input vector fields are identified with a log-normal process, our controller will automatically be cautious in scarcely explored regions.
	%
	%

\section{Method}
\label{sec:method_gpctrl_fullyactuated}
\subsection{Model}
\textbf{Dynamics.} Let $I\subset \Real$ be a (usually continuous) set of times, $\configspace$ denote the configuration space, $\statespace$ the state space and $\ctrlspace$ the control space.
Via the principle of least action and the resulting Euler-Lagrange equation, Lagrangian mechanics leads to the conclusion that controllable mechanical systems are of second order and can be written in \textit{control-affine }form:
\begin{equation}
\ddot q = a(q,\dot q) + b(q,\dot q) u.
\end{equation}

Here, $q \in \configspace$ is a generalized coordinate of the configuration and $u \in \ctrlspace$ is the control input.
Functions $a,b$ are called drift and input functions, respectively.
In the pendulum control domain we consider below, $q$ will encode joint angles and $u$ is a torque $\ddot q$ is proportional to.


Defining $x_1 := q$, $x_2 := \dot q \in \configspace$, we can write the state as $x := [x_1,x_2]$. 
The dynamics can be restated as the system of equations

\begin{align}
\dot \state_1 &= \state_2\\
\dot \state_2 &= a(\state_1,\state_2) + b(\state_1,\state_2) u \\
				 & = a(\state_1,\state_2) + \sum_{j=1}^m u_j \, b_j(\state_1,\state_2) 								\label{Eq:dyn2norderctrlaff}
\end{align}
where $m = \dim \ctrlspace$ and $b_j(\state_1,\state_2)$ is the $j$th row of matrix $b(\state_1,\state_2) \in \Real^{n \times m}.$
In this work, we assume the system is \textit{fully actuated}. That is, we assume that $b(q,\dot q)$ always is full-rank: $\rank b(q,\dot q)=\dim Q =:n, \forall q$. That is, full-actuation enables us to instantaneously set the acceleration in all dimensions of $\configspace$. However, we do not have immediate control over joint-angle velocities.
Incorporating this kind of knowledge afforded by Lagrangian mechanics is beneficial both from a principled Bayesian vantage point and in order to decompose the dimensionality of the learning task.

\textbf{Epistemic uncertainty and learning.} Both dynamics functions $a$ and $b$ can be uncertain a priori. That is, a priori our uncertainty is modelled by the assumption that $a \sim \Pi^a, b \sim \Pi^b$ where $ \Pi^a,\Pi^b$ are stochastic processes.
The processes reflect our epistemic uncertainty about the true underlying (deterministic) dynamics functions $a$ and $b$. If data becomes available over the course of the state evolution, we can update our beliefs over the dynamics in a Bayesian fashion. That is, at time $t \in I$ we assume $a \sim \Pi^a | \mathcal D_t, b \sim \Pi^b | \mathcal D_t$ where $\mathcal D_t$ is the data recorded up to time $t$. The process of conditioning is often referred to as (Bayesian) \emph{learning}.

\textbf{Data collection.} We assume our controller can be called at an ordered set of times $ I_u \subset I$. At each time $t \in I_u$, the controller is able to observe the state $\state_t = x(t)$ \footnote{In fact, we can only observe $q$ and have to obtain noisy observations of $\dot q$ as we will describe below.} and to set the control input $u_t = u(t,x_t)$. The controller may choose to evoke learning at an ordered subset $I_\lambda \subset I_u$ of times. To this end, at each time $\tau \in I_\lambda$, the controller evokes a procedure explicated in Sec. \ref{sec:learnproc} if it decides to incorporate an additional data point $(t,x_t,u_t)$ into data set $\mathcal D_t $  $(t >\tau)$. The decision on whether to update the data will be based on the belief over the data point's anticipated informativeness as approximated by its variance.\footnote{Variance is known to approximate entropic measures of uncertainty (cf. \cite{Alpcan2011}) and often easier to compute than entropy.}

For simplicity, we assume that learning can occur every $\Delta_\lambda$ seconds and the controller is called every $\Delta_u \leq \Delta_\lambda$ seconds. A continuous control takes place in the limit of infinitesimal $\Delta_u$.
\subsection{Learning procedure}
\label{sec:learnproc}
To enable learning, we will require derivatives of the state (that is estimates of $\ddot q$ and $\dot q$). If we do not have physical means to measure velocities and accelerations, obtaining numerical estimates becomes necessary based on observations of $q(t) = x_1(t)$. To estimate derivatives, we chose a second-order method. That is, our state derivative estimates are $\dot y(t_i + \Delta_o) := \frac{ x(t_i + 2\Delta_o) - x(t_i)}{2 \Delta_0} $ where $\Delta_o$ is a period length with which we can observe states. In this work, we assume $\Delta_o = \Delta_u$.

Assuming online learning, the data sets $\mathcal D_t$ are found incrementally. Since it is hard to use the data to infer $a$ and $b$ simultaneously, we will have to actively decide which one we desire to learn about (and set the control accordingly -- which we will then refer to as a \textit{probing control}). To this end, we distinguish between the following learning components:

\begin{itemize}

\item Learning $a(x)$: Assume we are at time $t \in I_\lambda$ and that we decide to learn about $a$. This decision is made, whenever our uncertainty about $a_t :=a(x_t)$, encoded by $\var {a(x_t)}$, is above a certain threshold $\theta_{\text{var}}^a$. When learning is initiated, we keep the control constant for two more time steps $t+\Delta_u$, $t+2\Delta_u$ to obtain a good derivative estimate as described above. To remove additional uncertainty due to ignorance about $b$, we set probing control $u_t=u_{t+\Delta_u} = u_{t+2 \Delta_u} =0$ yielding dynamics $\dot x_2 = a(x)$ during time interval $ [t,t+2 \Delta_u)$.
On the basis of a derivative estimate $\dot y_2(t)$, we can determine a noisy estimate $\tilde a_{t+\Delta_u}$ of unknown function value $a_{t+\Delta_u}$ at time $t$ as per 
\[ \tilde a_{t+\Delta_u} = \dot y_2({t+\Delta_u}). \] So, $({t+\Delta_u}, \tilde a_{t+\Delta_u}, 0)$ is added to the data after time ${t+2\Delta_u}$.

\item Learning $b_j(x)$: At time $t \in I_\lambda$, we choose to learn about function $b_j$ whenever our uncertainty about $a(x(t_i))$ is sufficiently small (i.e.  $\var {a(x_i)} \leq \theta_a$) and our uncertainty about $b_j$ is sufficiently large ($\var {b_j(x_i)} > \theta_b$). When learning is initiated, we keep the control constant for two more time steps $t+\Delta_u$, $t+2\Delta_u$ to obtain a good derivative estimate as described above.

Let $e_j \in \Real^m$ be the $j$th unit vector. To learn about $b_j(x)$ at state $x$, we apply a control action $u = u_j e_j$ where $u_j \in \Real \backslash\{0\}$. Inspecting Eq. \ref{Eq:dyn2norderctrlaff} we can then see that $b_j(x) = \frac{\dot x_2 -a(x)}{u_j}.$
Since $a(x)$ will generally be a random variable, so is $b_j(x)$ having mean $\expect{ b_j(x) }= \frac{ \dot x_2 - \expect{ a(x)}}{u_j}$ and variance $\var{ b_j(x) }= \frac{ 1 }{u_j^2} \var{ a(x)}$.
We obtain a noisy estimate $\dot y$ of its derivative analogously to above. Modelling $\dot x_2$ as a random variable with mean $\dot y_2$, $b_j(x)$ becomes a random variable with mean 
\begin{equation} \expect{ b_j(x) }= \frac{ \dot y_2 - \expect{ a(x)}}{u_j} \end{equation} and variance \begin{equation} \var{ b_j(x) }= \frac{\var{\dot x_2} +\var{ a(x)} }{u_j^2} \leq \frac{\var{\dot x_2} +\theta_a }{u_j^2}. \label{eq:varb_j} \end{equation}
Therefore, after time $t+ 2\Delta_u$, we add training point $\Bigl(x_{t+\Delta_u}, \expect{b_j(x_{t+\Delta_u})},u_{t}\Bigr)$ to the data set. The additional variance (as per Eq. \ref{eq:varb_j}) is captured by setting observational noise levels for $\Pi^b$ accordingly.
\end{itemize}

\subsection{Control law}
%
%
%

Unless the control actions are chosen to aid system identification (as described above), we will want to base our control actions on our probabilistic belief model over the dynamics. 
Given such an uncertain model, it remains to define an (outer-loop) control policy $u$ with desirable properties. In this work, we propose to define a control law that, when not learning, uses the probabilistic model to guarantee desired behaviour in expectation. In this work, our attention will be restricted to expected stability.


%


Let $a_t :=  a(x(t))$, $b_t:=b(x(t))$ and $q_t := q(t)$. Acceleration $\ddot q_t = a_t + b_t u(t)$ is a random variable with mean  $\expect{\ddot q_t} = \expect{a_t | \mathcal D_t} + \expect{b_t | \mathcal D_t} u.$

Hence, when applying inversion control law 

\begin{equation}\label{eq:fblin_law_means}
	u(t,x; u') :=  \expect{b(x) | \mathcal D_t}^{-1} \bigl( -\expect{ a(x) | \mathcal D_t} +u' \bigr)
\end{equation}

we get an expected closed-loop dynamics of 

\begin{align}
\expect{\dot q_t| \mathcal D_t} &= \expect{ x_2(t) | \mathcal D_t}=\expect{\dot x_1(t) | \mathcal D_t} =  \dot y_1 \label{eq:fblin_dyn_expected0}\\ 
	\expect{\ddot q_t | \mathcal D_t, x} & =\expect{\dot x_2(t) | \mathcal D_t,x}= \expect{a_t | \mathcal D_t} + \expect{b_t | \mathcal D_t}   \expect{b_t | \mathcal D_t}^{-1} \bigl( -\expect{ a_t | \mathcal D_t} +u' \bigr) = u'\\
		\expect{\dot x_2(t) | \mathcal D_t}& = \int_{\statespace} \expect{\dot x_2(t) | \mathcal D_t,x} \d P(x) = \int_{\statespace} u'(t,x) \d P(x) = \expect{u'(t,x)}. 	\label{eq:fblin_dyn_expected}
\end{align}

where $u'(t,x)$ is an \emph{inner-loop} control law. 

\begin{thm}
Assume we are not performing probing actions anymore. That is, we are at time $t \geq t_0$ such that $t_0 > \sup I_\lambda$. Let $u'(t,x)$ be a control law that is linear in state $x$ and that ensures the double-integrator dynamics of the form
\[ 	\dot z_1 = z_2,
	 \dot z_2 =u'(t,z;\xi) \]
to have $\xi$ as a globally asymptotically stable equilibrium point. Finally, suppose that expectation and derivative commute. That is, $\nabla_t \expect{ x_i(t) | \mathcal D_{t_0}} = \expect{\nabla_t x_i(t) | \mathcal D_{t_0}}$.
Then, our control law as per Eq. \ref{eq:fblin_law_means}, with inner control law $u'(x,t;\xi )$, ensures $\xi$ is a globally asymptotically stable equilibrium of the expected dynamics.  In particular, $\lim_{t \to \infty}\norm{\expect{ q_t - \xi_1 | \mathcal D_{t_0}} }^2 =  0 \wedge \lim_{t \to \infty}\norm{\expect{ \dot q_t - \xi_2 | \mathcal D_{t_0}} }^2 =  0$.

\begin{proof} (Sketch)  
Let $\nabla_t$ denote  the differential operator with respect to time. Leveraging the linearity of the differential operator, we can exchange it with the expectation operator. Thereby, we conclude from Eq. \ref{eq:fblin_dyn_expected0}  and  Eq. \ref{eq:fblin_dyn_expected} that

$ \nabla_t \expect{ x_1(t) | \mathcal D_{t_0}} = \expect{\nabla_t x_1(t) | \mathcal D_{t_0}} = \expect{ x_2(t) | \mathcal D_{t_0}}$ and  

$\nabla_t \expect{ x_2(t) | \mathcal D_{t_0}} = \expect{ \nabla_t  x_2(t) | \mathcal D_{t_0}} = \expect{ u'(t,x; \xi)} =
 u'(t,\expect{x};\xi)$ where the last step follows by linearity of the control law.
%
Defining $z_i := \expect{ x_i(t) | \mathcal D_{t_0}} $ yields the quadratic regulator problem :
$ \nabla_t z_1 = z_2, \nabla_t z_2 =  u'(t,z)$. By assumption, we know that $u'$ ensures that $\xi$ is a globally asymptotic equilibrium point of this dynamic system. Hence,  
in particular, $\lim_{t \to \infty}\norm{ z_1(t) - \xi_1 }^2 =  0 \wedge \lim_{t \to \infty}\norm{ \dot z_2(t) - \xi_2  }^2 =  0$. Resubstituting the definitions of $\expect{ x_i(t) | \mathcal D_{t_0}} $ for $z_i$ and subsequently, of $q = x_1, \dot q = x_2$, yields the desired statement.
%
%
%
%
\end{proof}
\end{thm}

\begin{rem}
The requirement that differential and expectation operator can be interchanged has to be checked on a case by case basis and 
depends on the interpretation of the random differential equation or the particular kinds of processes driving the equation.
Examples of where the assumption is met are white-noise limits or, when the integral curve $x(t)$ is an $L_2$ process with differentiable covariance function (here, $\nabla_t x$ is the mean-square derivative). An alternative would be to show expected stability for every Euler-approximation and prove convergence in the mean of these Euler approximations to the continuous-time equation.  
\end{rem}

Consequently, we have given conditions under which  our control law guarantees feedback-linearisation in expectation (and of the dynamics of the mean trajectory). That is, by choosing $u'$ to impose desired behaviour for the double integrator problem $\ddot q = u'$ (which is easy), we can re-shape the dynamics such that the expected closed-loop dynamics is stable. 

For instance, a simple method of guaranteeing global asymptotic convergence of the state towards a goal state $\xi = [\xi_1,\xi_2]$ would be to set the inner-most control law to the proportional feedback law

\begin{equation} \label{eq:innerpropctrl} u' (t,x ;w) := w_1 (\xi_1 - x_1)  + w_2 (\xi_2 - x_2) \end{equation} 

where $w_1,w_2 >0$.

\section{Experiments -- Learning to control a torque-controlled damped pendulum with a combination of normal and log-normal processes} \label{sec:experiments}
We explored our method's properties in simulations of a rigid pendulum with (a priori unknown) drift $a(x) :=  - \frac{g}{l} \sin (x_1) - \frac{r(x_1)}{m l^2} x_2$ and constant input function $b(x) = \frac{1}{m \, l^2}$.  Here, $x_1 = q, x_2 = \dot q \in \Real$ are joint angle position and velocity, $r$ denotes a friction coefficient, $g$ is acceleration due to gravity $l$ is the length and $m$ the mass of the pendulum.  The control input $u \in \Real$ applies a torque to the joint that corresponds to joint-angle acceleration.  The pendulum could be controlled by application of a torque $u$ to its pivotal point. $q=0$ encode the pendulum pointing downward and $q=0$ denoted the position in which the pendulum is upward. Given an initial configuration $x_0 = [q_0, \dot q_0]$ we desired to steer the state to a terminal configuration $\xi = [q_f, 0]$.
            

For learning, we assumed that $a \sim \GP(0,K_a)$ and $b \sim \log \GP(0,K_b)$ had been drawn from a log-normal process.\footnote{For details on normal processes see \cite{GPbook:2006}.} The latter assumption encodes a priori knowledge that control input function $b$ can only assume positive values (but, to demonstrate the idea of cascading processes, we had discarded the information that $b$ was a constant). During learning, the latter process was based on a standard normal process conditioned on $\log$-observations of $\tilde b$.
To compute the control as per Eq. \ref{eq:fblin_law_means}, we need to  convert the posterior mean over $\log b$ into the expected value over $b$. The required relationship is known to be as follows: 
\begin{equation} \label{eq:postmean_lognormal}\expect{b(x) | \mathcal D_t} =  \exp\Bigl( \expect{ \log b(x) | \mathcal D_t} + \frac{1}{2}\var{\log b(x) | \mathcal D_t} \Bigr).
\end{equation}
If required the posterior variance can be obtained as 
\[\var{b(x) | \mathcal D_t} =  \exp\Bigl( 2 \expect{ \log b(x) | \mathcal D_t} + \var{\log b(x) | \mathcal D_t} \Bigr)  \exp\Bigl(  \var{\log b(x) | \mathcal D_t} - 1 \Bigr).\]

Note, the posterior mean over $b$ increases with the variance of our normal process in log-space, and, the control  law as per Eq. \ref{eq:fblin_law_means} is inversely proportional to the magnitude of this mean. Hence, the resulting controller is \textit{cautious}, in the sense that control output magnitude is damped exponentially in regions of high uncertainty (variance).


To simulate a discrete  $0$th order sample-and-hold controller in a continuous environment, we simulated the dynamics between two consecutive controller calls (occurring every $\Delta_u$ seconds) employing standard ODE-solving packages (i.e. Matlab's ode45 routine). 


%
We illustrated the behaviour of our controllers in a sequence of four experiments. The parameter settings are provided in Tab. \ref{Tab:pars}. Recorded control energies and errors (in comparison to continuous proportional controllers) are provided in Tab. \ref{Tab:resenganderr}. \\
Our Bayesian controller maintains an epistemic beliefs over the dynamics. These beliefs govern our control decisions (including those when to learn).  Furthermore, to keep prediction times low, beliefs are only updated when the current variance indicated a sufficient of uncertainty. Therefore, one would expect to observe three properties of our controller:

(i) When the priors are chosen sensibly (could be indicated by the dynamic functions' likelihood under the probabilistic models), we expect good control performance.

(ii) Prior training improves control performance and, reduces learning, but is not necessary to reach the goal.
Both properties can be observed in Exp.1 and Exp. 3.

(iii) When the controller is ignorant of the inaccuracy of its beliefs over the dynamics (i.e. the actual dynamics are unlikely but the variances are low), control may fail since the false beliefs are not updated. An example of this is provided in Exp. 2.

(iv) We can overcome such problems practically, by employing the standard technique (see \cite{GPbook:2006}) of choose the prior hyper-parameters that maximise marginal likelihood. In Exp. 3, this approach was successfully applied to the control problem of Exp. 2.


 \begin{table}[hbt!] 
 \centering
 \begin{footnotesize} 
 \begin{tabular}{@{}lccccccccc@{}}
 \toprule
 \cmidrule(r){2-4} \cmidrule(l){5-9} 
 $Parameter(s):$ & (l,r,m) &  $\Delta_u$&  $\Delta_l$&  $(\theta_{var}^a,\theta_{var}^{\log \, b})$ &$x_0$& $\xi$ & $(w_1,w_2)$&$t_f$ \\
 \midrule

\textit{Exp. 1} & (1,1,0.5) 	& .01 & .5	& (.001, .005) 	& (0,-2) & ($\pi,0$) & (1,1)&20\\
\textit{Exp. 2} & (1,0.5,4) 	& .01 & 1	& (.001, .005) 	& (0,-2) & ($\pi,0$) & (2,2)&15\\
\textit{Exp. 3} & (1,0.5,4) 	& .01 & 1	& (.001, .005) 	& (0,-2) & ($\pi,0$) & (2,2)&20\\
 \bottomrule
 \end{tabular}
 \end{footnotesize} 
 \caption{Parameter settings.}   
\label{Tab:pars}
 \end{table}



 \begin{table*}[hbt!]  
 \centering
 \begin{footnotesize} 
 \begin{tabular}{@{}lcccccccccc@{}}
 \toprule
 & \multicolumn{3}{c}{$\int_I u_{adapt}^2(t) dt$ } & 
 \multicolumn{5}{c}{$\int_I (x(t)- \xi)^2 dt$}  & 
 \multicolumn{2}{c}{$(\abs{\mathcal D^a_{t_f}},\abs{\mathcal D^b_{t_f}})$}\\
 \cmidrule(r){2-5} \cmidrule(l){6-9}  \cmidrule(l){10-11}
 $Controller:$ & P1 &P100 & SP1 & SP2 &P1 &P100 & SP1 & SP2 & SP1 & SP2 \\
 \midrule
  \textit{Exp. 1 }		&134 & 644 & 139&  57& 137& 10 &  59&   25&   (18, 20)   &(23, 53)\\
	\textit{Exp. 2} &  552   & 11942  &  14759 &   17029   & 139& 10 &   82    & 72   & (2,1) & (2,1)\\
	\textit{Exp. 3} &  730  &  11942  &  3753   & 1619 & 184 & 10 & 83  &  17   &(12,2) & (12,2)\\
 \bottomrule
 \end{tabular}
 \end{footnotesize}
  \caption{Cumulative control energies, squared errors and data sizes (rounded to integer values). P\emph k: P-controller with feedback gain $k$. P1 failed to reach the goal state in all experiments. High-gain controller P100 succeeded in reaching the goal in all experiments but required a lot of energy. SP1: stochastic process -based controller with empty data set to start with. SP2: reset SP1 with training data collected from the first run. } 
	\label{Tab:resenganderr}
 \end{table*}


\textbf{Experiment 1.}
We started with a zero-mean normal process prior over $a(\cdot)$ endowed with a rational quadratic kernel with automated relevance detection (RQ-ARD) \cite{GPbook:2006}. The kernel hyper-parameters were fixed. Observational noise variance was set to $0.01$.
The log-normal process over $b(\cdot)$ was implemented by placing a normal over $\log b(\cdot)$ with zero mean and RQ-ARD kernel with fixed hyper-parameters and observational noise level $0.1$. Note, the latter was set higher to reflect the uncertainty due to $\Pi^a$. In the future, we will consider incorporating hetereoscedastic observational noise based on $\var a$ and the sampling rate. Also, one could incorporate knowledge about periodicity in the kernel.

\begin{figure}
\centering
\begin{subfigure}{.49 \textwidth}
  \centering
  \includegraphics[width = 3.5cm, clip, trim = 3.5cm 9.5cm 4.5cm 9cm]{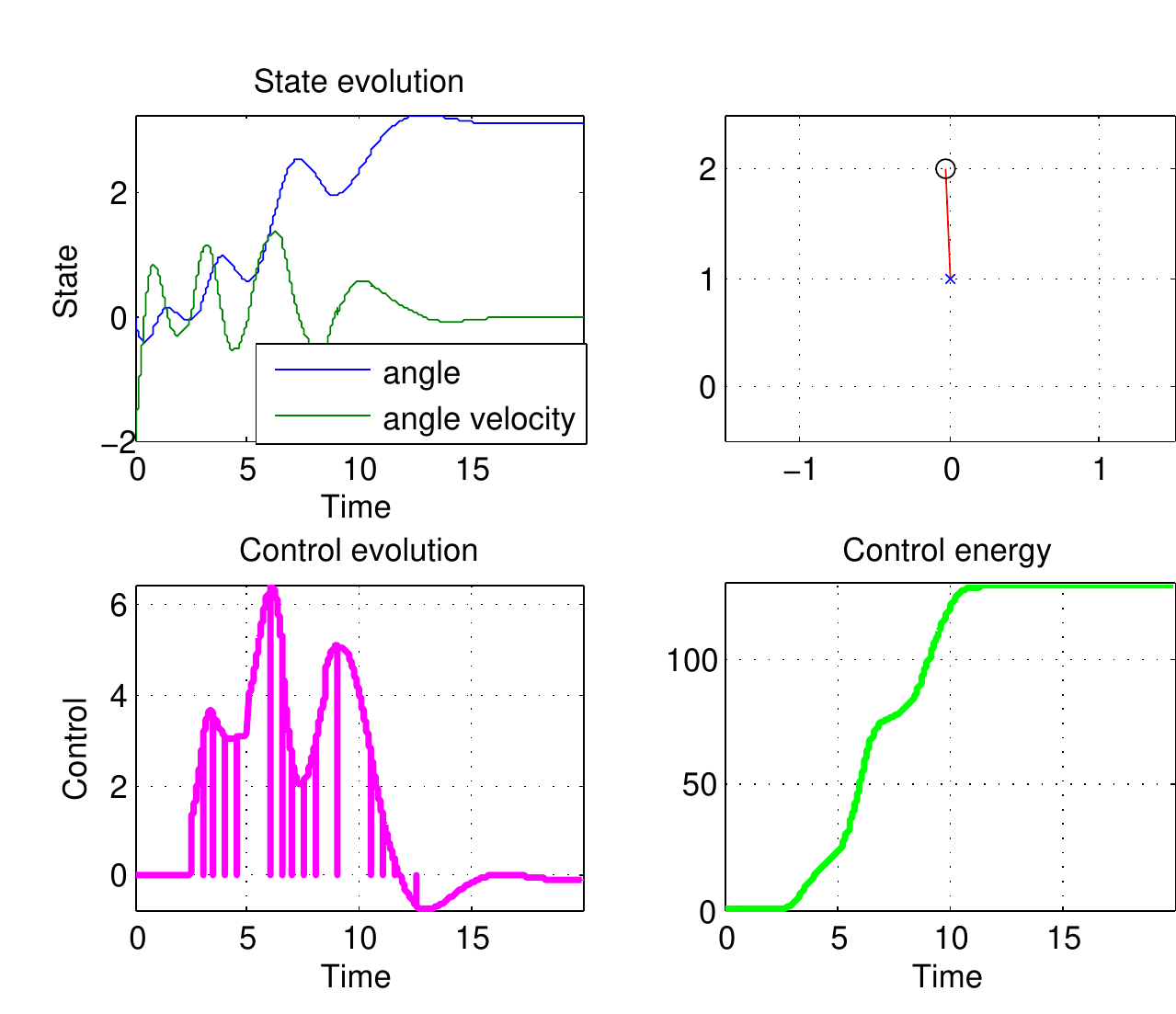}
  \caption{Control with untrained prior.}
  \label{fig:sub1exp1}
\end{subfigure}%
\begin{subfigure}{.48\textwidth}
  \centering
  \includegraphics[width = 3.5cm, clip, trim = 3.5cm 9.5cm 4.5cm 9cm]{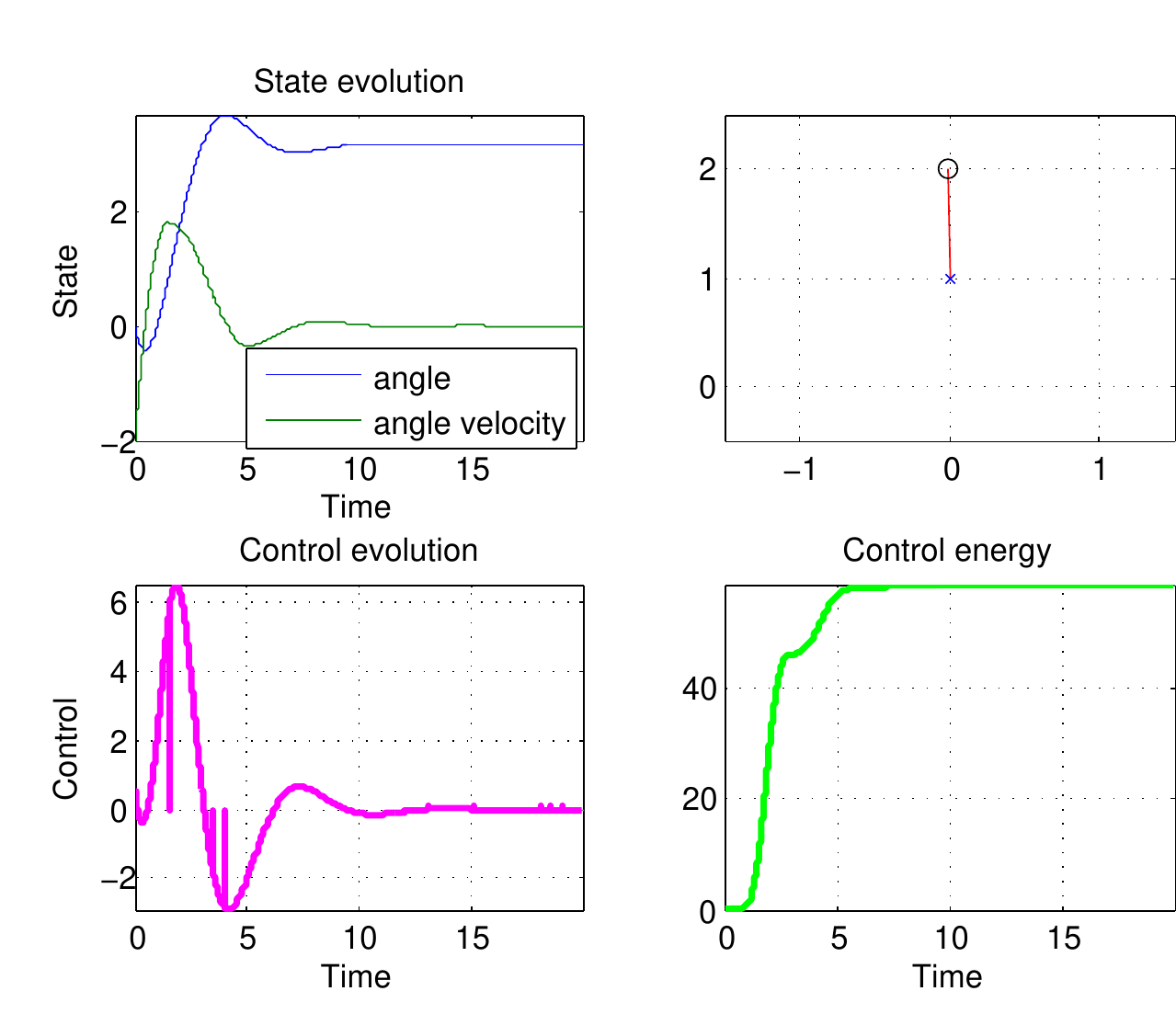}
  \caption{Control evolution with trained prior from the first round.}
  \label{fig:sub2exp1}
\end{subfigure}
\caption{Experiment 1. Comparison of runs with untrained and pre-trained processes. The top-right image shows the final position of the pendulum having successfully reached the target angle $\xi_1=\pi$.  The dips in the control signal represent probing control actions arising during online learning.}
\label{fig:exp1}
\end{figure}

\begin{figure}
\centering
\begin{subfigure}{.3\textwidth}
  \centering
  \includegraphics[width=1\linewidth]{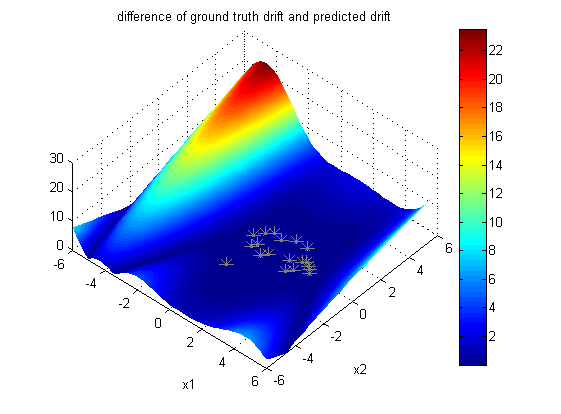}
  	 \caption{$\abs{a(x) - \expect{a(x) | D_{t_f}}}$.}
  \label{fig:sub6exp1}
\end{subfigure}
\begin{subfigure}{.3\textwidth}
  \centering
  \includegraphics[width = 3cm, clip, trim = 1cm 0cm 1cm 1cm]
	{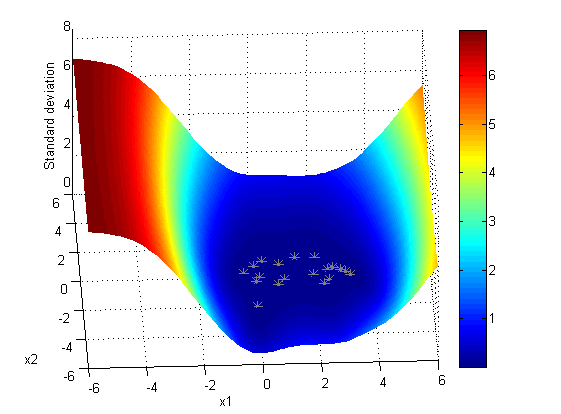}
	\caption{Posterior standard deviation $\sqrt{\var{a(x) | D_{t_f}}}$.}
  \label{fig:sub5exp1}
\end{subfigure}
\begin{subfigure}{.33\textwidth}
  \centering
  \includegraphics[width = 3cm, clip, trim = 1cm 0cm 1cm 1cm]
	{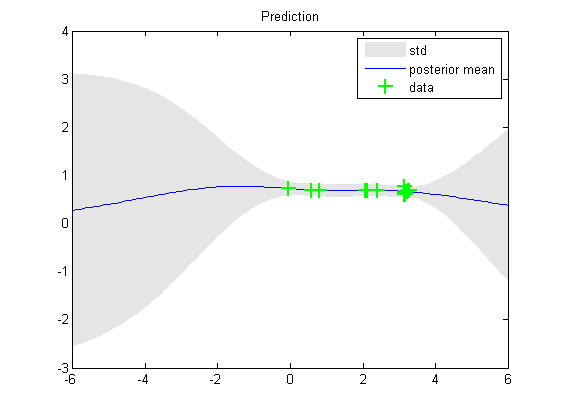}
\caption{Posterior over $\log b$.}
  \label{fig:sub4exp1}
\end{subfigure}
\caption{Experiment 1. Posterior models of SP1. Stars indicate training examples. The stochastic process has learned the dynamics functions in explored state space with sufficient accuracy.}
\label{fig:exp1_2}
\end{figure}

Results are depicted in Fig. \ref{fig:exp1} and \ref{fig:exp1_2}.
We see that the system was accurately identified by the stochastic processes. When restarting the control task with stochastic processes pre-trained from the first round, the task was solved with less learning, more swiftly and with less control energy. 

\textbf{Experiment 2.} \label{exp3bayesctrl1} 
%
  %
We investigated the impact of inappropriate magnitudes of confidence in a wrong model. We endowed the controller's priors with zero mean functions and SE-ARD kernels \cite{GPbook:2006}. 
Length scales of kernel $K_a$ were set to 20 and the output scale to $0.5$. In addition to the low output-scale, we set observational noise variance to a low value of 0.0001 suggesting (ill-founded) high confidence in the prior.
The length scale of kernel $K_b$ was set to 50 with low output scales and observational noise variance of 0.5 and 0.001, respectively.

\begin{figure}
\centering
\begin{subfigure}{.49\textwidth}
  \centering
  \includegraphics[width = 3cm, clip, trim = 4.5cm 9.5cm 4.5cm 9cm]{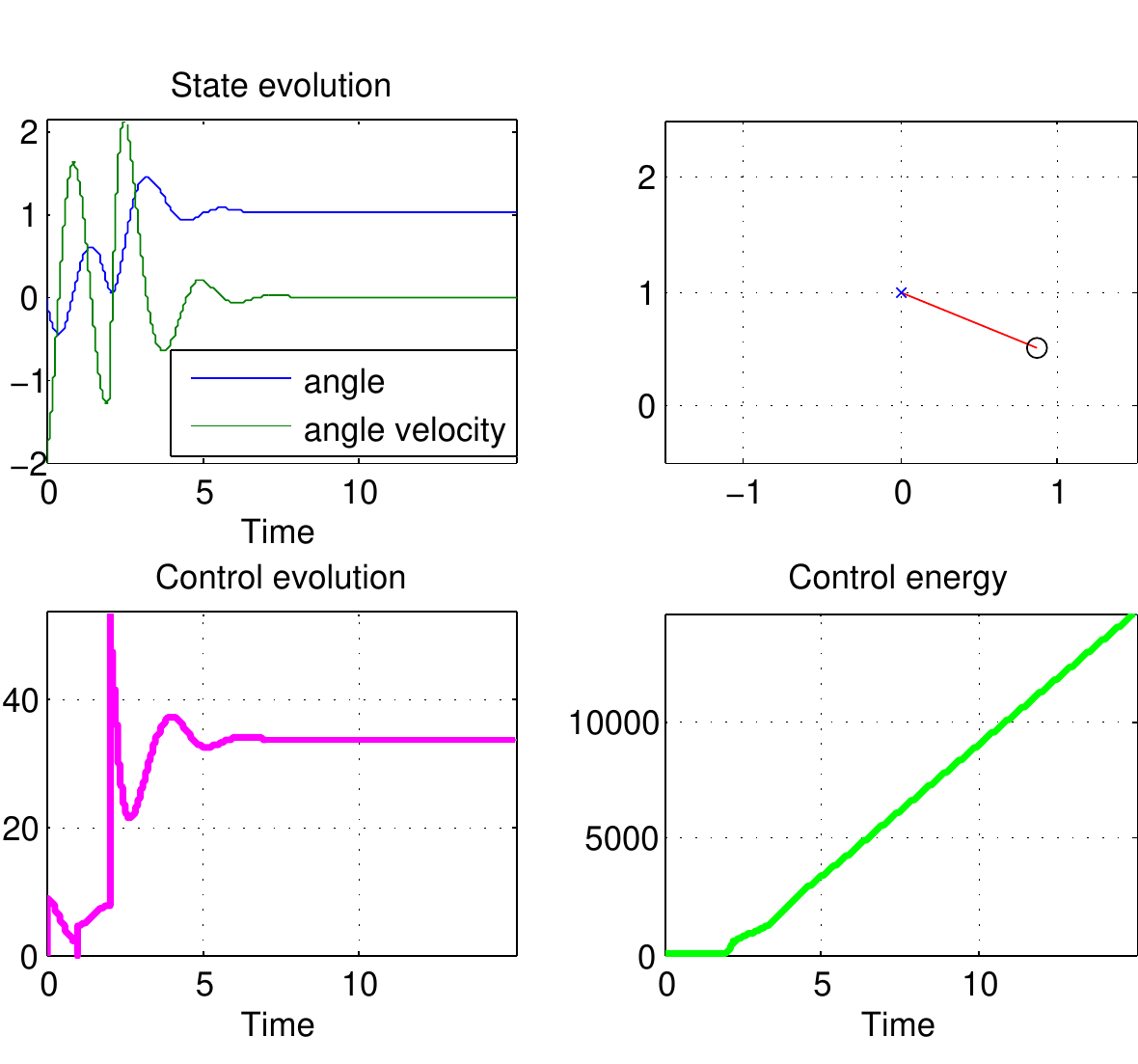}
  \caption{Control evolution of SP1, with untrained prior.}
  \label{fig:sub1exp3}
\end{subfigure}%
\begin{subfigure}{.49\textwidth}
  \centering
  \includegraphics[width = 3cm, clip, trim = 4.5cm 9.5cm 4.5cm 9cm]{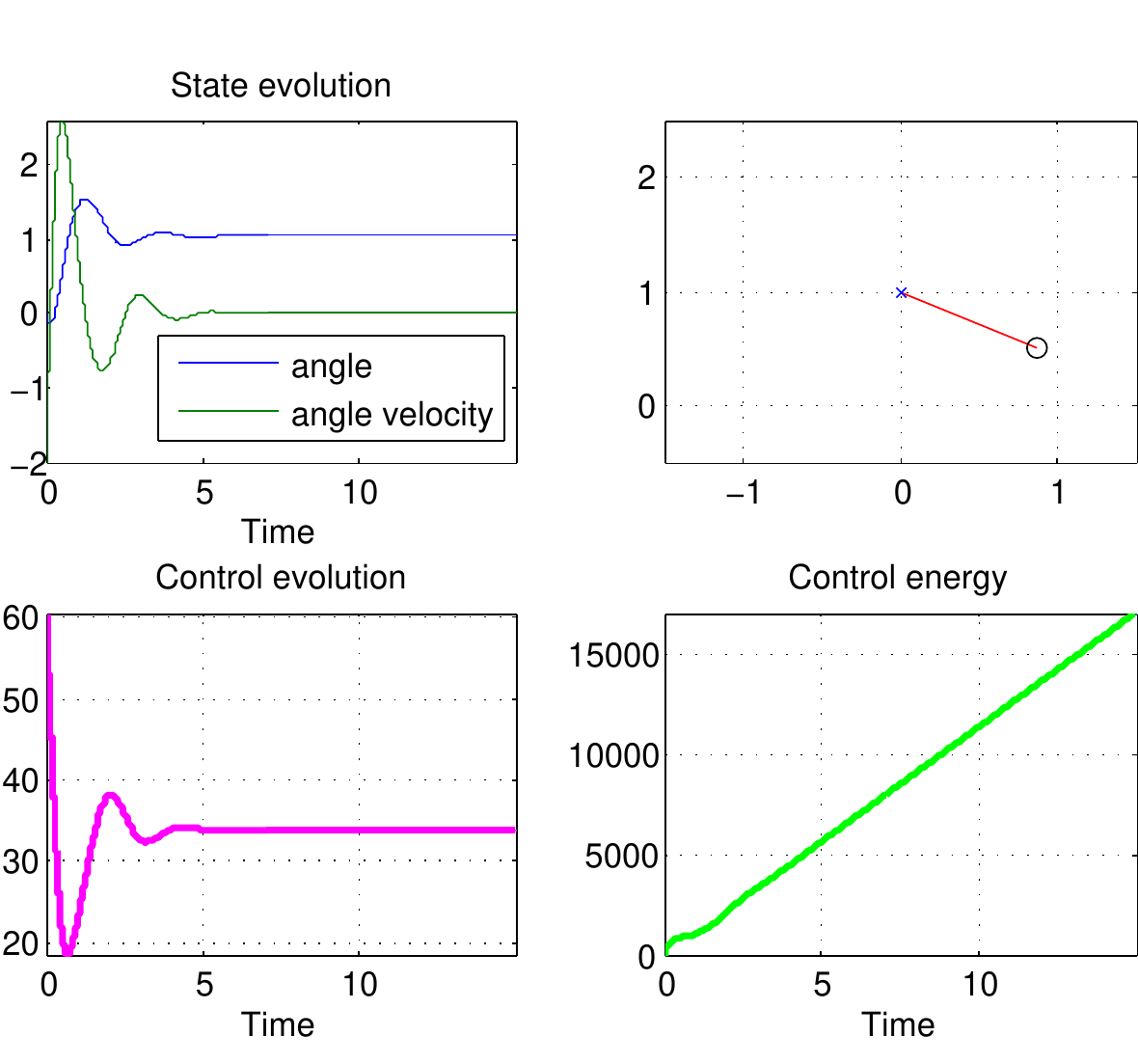}
  \caption{Control evolution of controller SP2, benefiting from learning experience from the first round.}
  \label{fig:sub2exp3}
\end{subfigure}
\caption{Experiment 2. Comparison of runs with untrained and pre-trained processes. Neither run succeeds in arriving at the target state due to being overly confident. }
\label{fig:exp3}
\end{figure}

The results, depicted in Fig. \ref{fig:exp3}. As to be expected, the controller fails to realise the inadequateness its beliefs. This results in a failure to update its beliefs and consequently, in a failure to converge to the target state.

Of course, this could be overcome with an actor-critic approach. Such solutions will be investigated in the context of future work.

\begin{figure}
\centering
%
\begin{subfigure}{.3\textwidth}
  \centering
  \includegraphics[width=1\linewidth]{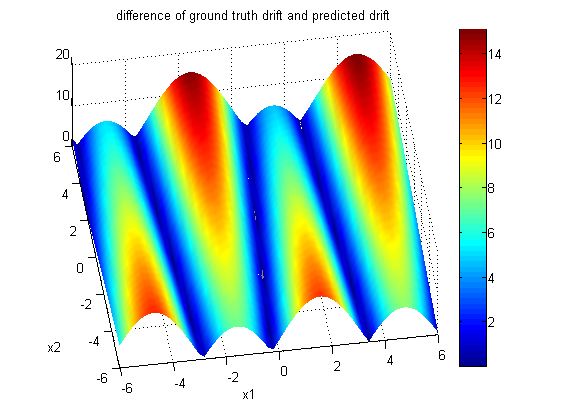}
  	 \caption{$\abs{a(x) - \expect{a(x) | D_{t_f}}}$.}
  \label{fig:sub6exp3}
\end{subfigure}
\begin{subfigure}{.33\textwidth}
  \centering
  \includegraphics[width=1\linewidth]{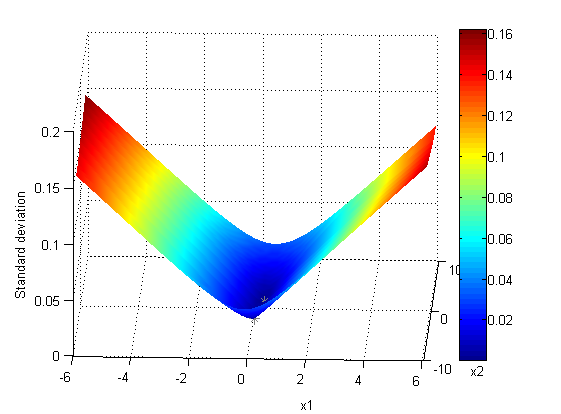}
	\caption{Posterior standard deviation $\sqrt{\var{a(x) | D_{t_f}}}$.}
  	  \label{fig:sub5exp3}
\end{subfigure}
\begin{subfigure}{.33\textwidth}
  \centering
  \includegraphics[width = 4cm, clip, trim = 1cm 4.5cm 1cm 4cm]{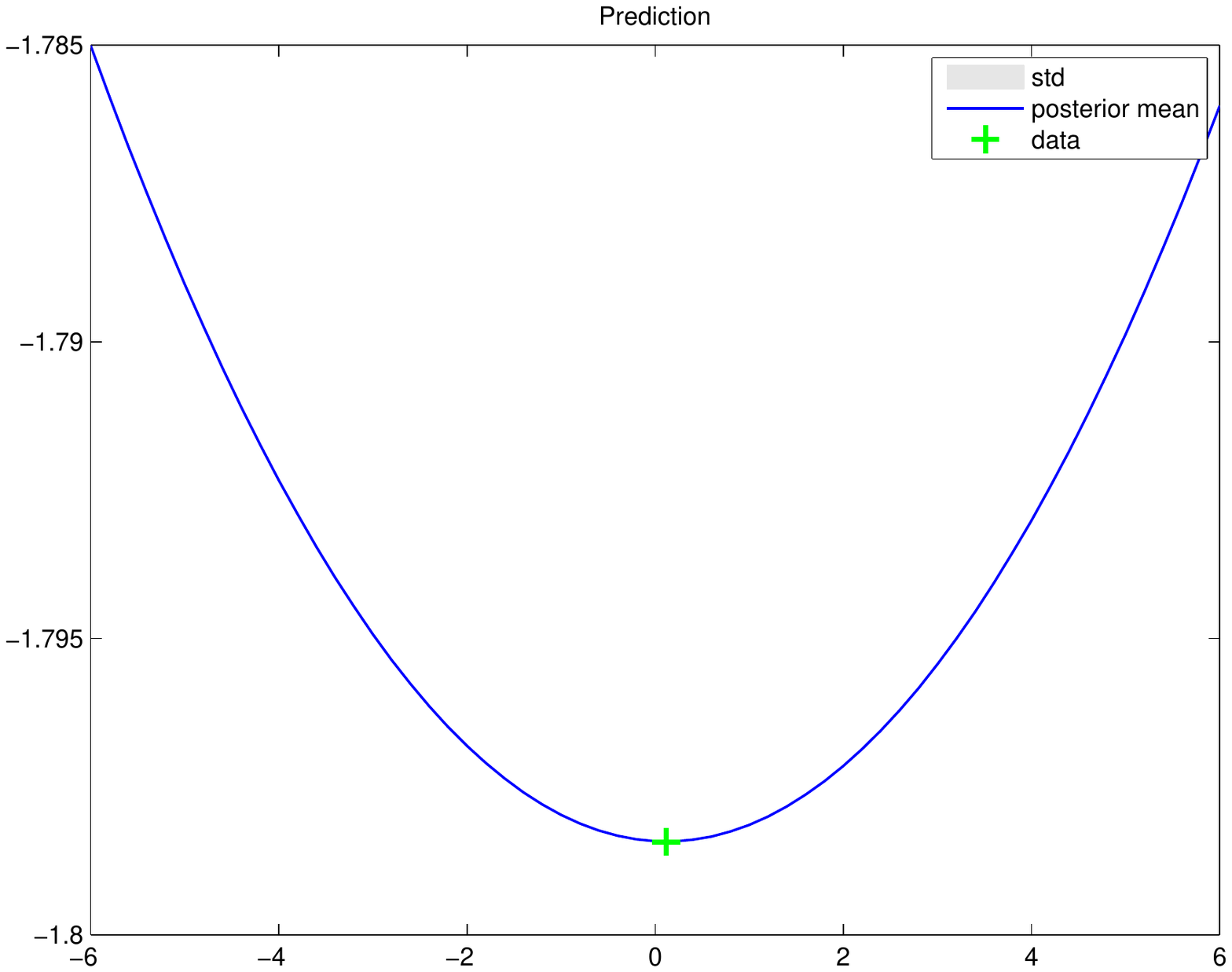}
\caption{Posterior over $\log b$.}
  \label{fig:sub4exp3}
\end{subfigure}
\caption{Experiment 2. Posterior models of SP1. Stars indicate training examples. Note, the low posterior variance suggests misleading confidence in an inaccurate model.}
\label{fig:exp3_2}
\end{figure}
\textbf{Experiment 3.}

Exp. 2 was repeated. This time, however, the kernel hyper-parameters were found by maximizing the marginal likelihood of the data. The automated identification of hyper-parameters is beneficial in practical scenarios where definition of a good prior for the underlying dynamics may be hard to conceive.

The optimiser succeeded in finding sensible parameters that allowed good control performance. As before, the method benefited from prior training yielding faster convergence and lower control effort. 
Both untrained and pre-trained methods outperformed the $P$-controllers either in terms of control energy or convergence. Finally, the SP controllers with hyper-parameter optimisation outperformed the SP controllers with fixed hyper-parameters set in Exp. 2 (c.f. Tab. \ref{Tab:resenganderr}).

\begin{figure}
\centering
\begin{subfigure}{.4 \textwidth}
  \centering
  \includegraphics[width = 3.5cm, clip, trim = 3.5cm 9.5cm 4.5cm 9cm]{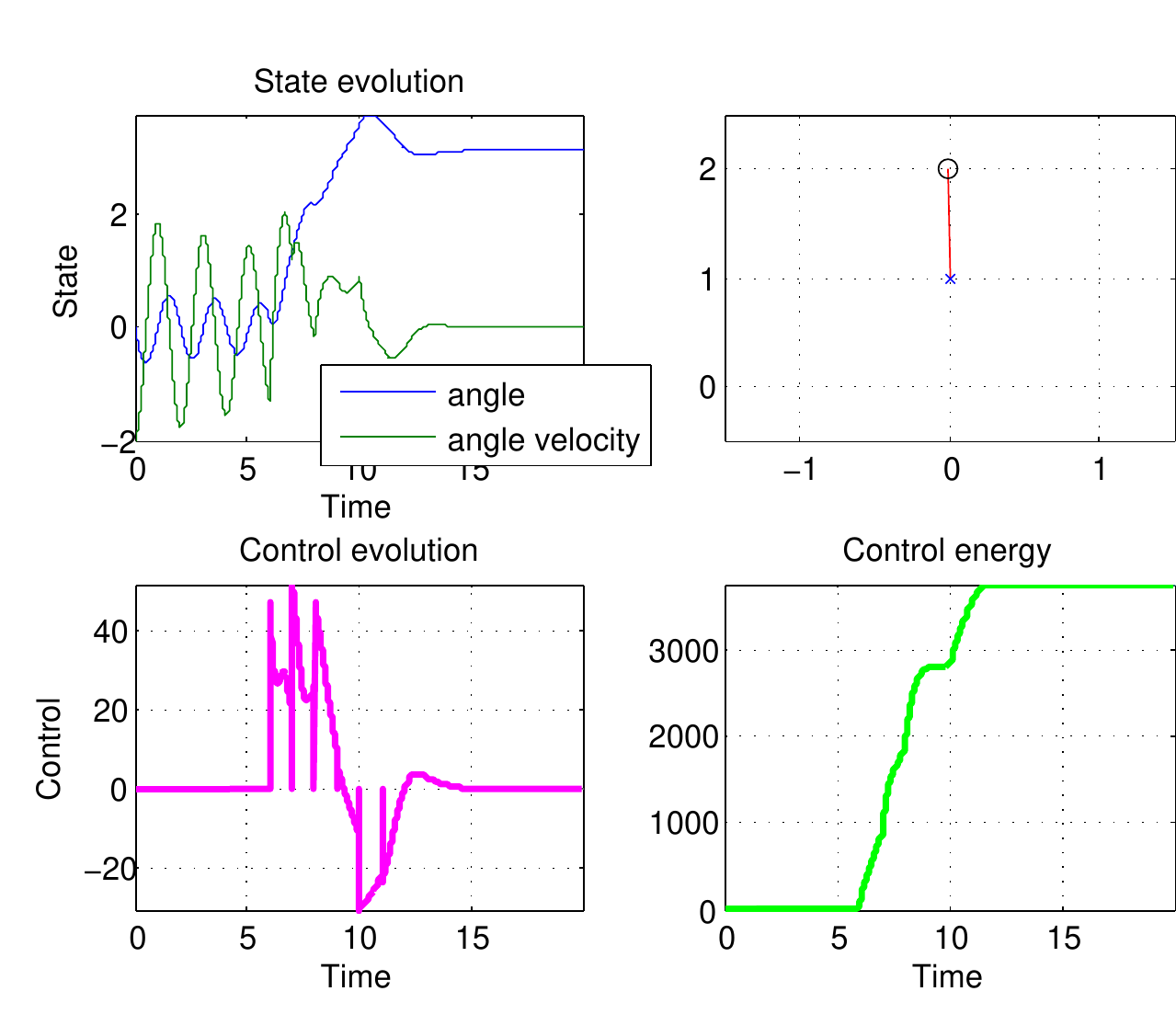}
  \caption{Control evolution of SP1, with untrained prior.}
  \label{fig:sub1exp4}
\end{subfigure}%
%
%
%
\begin{subfigure}{.26\textwidth}
  \centering
  \includegraphics[width=1\linewidth]{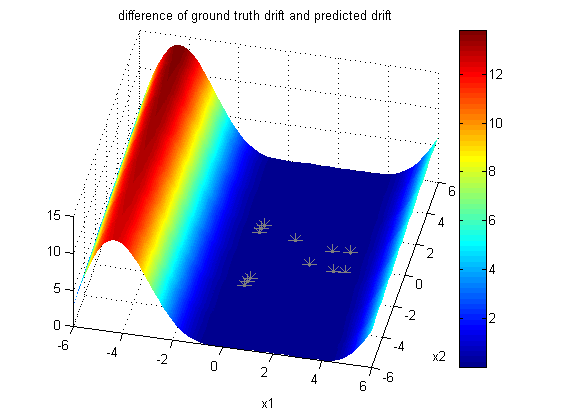}
  	 \caption{$\abs{a(x) - \expect{a(x) | D_{t_f}}}$.}
  \label{fig:sub6exp4}
\end{subfigure}
\begin{subfigure}{.26\textwidth}
  \centering
  \includegraphics[width=1\linewidth]{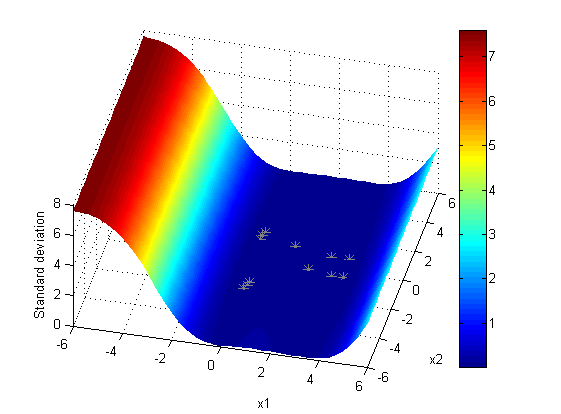}
	\caption{Posterior standard deviation $\sqrt{\var{a(x) | D_{t_f}}}$.}
  	
  \label{fig:sub5exp4}
\end{subfigure}
%
%
%

%
\caption{Experiment 3. Posterior models of SP1. Stars indicate training examples. The optimisation process succeeded in finding a sufficiently appropriate model.}
\label{fig:exp4_2}
\end{figure}

\section{Conclusions} We have applied Bayesian nonparametric methods to learn online the drift and control input functions of a fully-actuated control-affine second-order dynamical system. Paired with the idea of feedback-linearisation we devised a control law that switches between probing actions for learning and control signals that drive the expected trajectory towards a given setpoint. 
Our simulations have illustrated our controller's behaviour in the context of a pendulum regulator problem and that it can successfully solve the identification and control problems. They have also served as an illustration of the inherent pitfalls of Bayesian control -- that is, guarantees are stated relative to epistemological beliefs (encoded by a posterior) over the dynamical system in question. Therefore, the controller's performance may be undermined by ignorance over the potential falsity of prior beliefs (cf. Exp. 3). However, as illustrated in Exp. 3, even the most simple model selection methods can alleviate the burden of having to conceive a good fixed prior. 

In \emph{future work}, we will explore how to employ a predictor-corrector approach to uncover over-confidence of our models and to initiate learning. 
At present, our control law achieves desired performance of the expected trajectory. We will investigate how to extend the guarantees to achieve performance guarantees in expectation and within probability bounds. Other theoretical questions under investigation are analysis of the trade-offs between the impact of probing actions (to learn), the desire to keep prediction time low, information gain and control refresh cycle length $\Delta_u$. 
Finally, we will assess our methods' performance in higher-dimensional systems.

\begin{small}
\bibliographystyle{plain}

\end{small}

\end{document}